\documentclass[10pt,twocolumn,letterpaper]{article}

\pdfoutput=1

\usepackage{iccv}
\usepackage{times}
\usepackage{epsfig}
\usepackage{graphicx}
\usepackage{amsmath}
\usepackage{amssymb}
\usepackage{siunitx}
\usepackage{multirow} % for multirow table
\usepackage{adjustbox} % to resize table
\usepackage{balance}

\usepackage[pagebackref=true,breaklinks=true,letterpaper=true,colorlinks,bookmarks=false]{hyperref}

\iccvfinalcopy % *** Uncomment this line for the final submission

\def\iccvPaperID{25} % *** Enter the ICCV Paper ID here

\ificcvfinal\fi
\begin{document}

%%%%%%%%% TITLE
\title{Leaf Counting with Deep Convolutional and Deconvolutional Networks}

\author{Shubhra Aich and Ian Stavness \\
Computer Science, University of Saskatchewan\\
Saskatoon, Canada\\
{\tt\small s.aich@usask.ca, ian.stavness@usask.ca}
%\and
}

\maketitle
%\thispagestyle{empty}

%%%%%%%%% ABSTRACT
\begin{abstract}
In this paper, we investigate the problem of counting rosette leaves from an RGB image, an important task in plant phenotyping. We propose a data-driven approach for this task generalized over different plant species and imaging setups. To accomplish this task, we use state-of-the-art deep learning architectures: a deconvolutional network for initial segmentation and a convolutional network for leaf counting. Evaluation is performed on the leaf counting challenge dataset at CVPPP-2017. Despite the small number of training samples in this dataset, as compared to typical deep learning image sets, we obtain satisfactory performance on segmenting leaves from the background as a whole and counting the number of leaves using simple data augmentation strategies. Comparative analysis is provided against methods evaluated on the previous competition datasets. Our framework achieves mean and standard deviation of absolute count difference of $1.62$ and $2.30$ averaged over all five test datasets.
\end{abstract}

%%%%%%%%% BODY TEXT
\section{Introduction}
\label{intro}
Traditional plant phenotyping, which involves manual measurement of plant traits, is a slow, tedious and expensive task. In most cases, manual measurement techniques use sparse random sampling followed by the projection of those random measurements over the whole population which might incorporate measurement bias. Further, plant phenotyping has been identified as the current bottleneck in modern plant breeding and research programs \cite{furbank}.
%
% Furbank, R. T. and Tester, M. (2011). Phenomics?technologies to relieve the phenotyping bottleneck. Trends in plant science 16, 635?644
%
Therefore, interest in image-based phenotyping techniques have expanded rapidly over the past 5 years. Automation of the estimation of these visual traits up to a satisfactory level of accuracy using suitable computer vision techniques can boost production speed and reduce costs since fewer field technicians would be required for manual measurement each year.

In this paper, we work on estimating the number of leaves on a plant at the rosette stage, which is an indicator of plant health \cite{dataset-cvppp2017-01}.
%
% Minervini, M., Fischbach, A., Scharr, H., and Tsaftaris, S. A. (2015). Finely-grained annotated datasets for image-based plant phenotyping. Pattern Recognition Letters doi:http://dx.doi.org/10.1016/j.patrec.2015. 10.013
%
Our main objective is not only to develop a robust computer vision model, but also to generalize it so that the plant breeders can use this framework regardless of the plant species they are working on and of the quality of the image data they have acquired. Like one of the previous works \cite{cvppp2015_winner}, we also pose this problem as a nonlinear regression problem, where given the images, our framework approximates the count directly without segmenting individual leaf instances. This regression hypothesis is useful for a couple of reasons. First, although this nonlinear regression problem appears to be very high dimensional, it is usually more efficient than counting by identifying the individual leaf instances. Second, from the perspective of supervised machine learning, collecting ground-truth leaf counts is much simpler than generating ground-truth segmented regions for each leaf in the color images. In section \ref{experiments} of this paper, we show that the performance of the systems developed under the regression hypothesis is comparable to the state-of-the-art counting by instance segmentation approaches. However, unlike \cite{cvppp2015_winner}, we develop each of the components of our complete model in such a way that it can directly learn from the data without the need for manual heuristics or explicit knowledge on the plant species or other environmental factors. According to the state-of-the-art computer vision and machine learning literature, the best way to develop a generalized model without such prior knowledge is to use deep learning and therefore we adopt this paradigm in our work.

Similar to \cite{dpp-jubbens}, we train a deep convolutional neural network to count leaves by regression. However, the focus of our present work is to develop a single network that can generalize across different rosette datasets, rather than separate networks each built and tuned to maximize performance on an individual dataset. We also develop a deep convolutional-deconvolutional neural network for automatic whole plant segmentation and explore the effect of using a binary segmentation mask as an additional input channel to the leaf counting network in order to improve generalized performance. We evaluate our method as part of the Leaf Counting Challenge 2017 (LCC-2017) and report performance across the five subsets of the competition dataset. Through this work, we hope to inaugurate the research and development of a useful and generalized system for plant breeders to study leaf development in individual plants and eventually to study crop emergence in the field.

%Even though our method presented in this paper is engineered to participate in the Leaf Counting Challenge 2017 (LCC-2017) at the CVPPP-ICCV-2017 workshop, our motive is not to gain merely the highest possible accuracy over the specific dataset provided during the competition, but also to inaugurate the research and development of a useful and generalized system for the plant breeders. Thus, while it is likely possible to customize and tune our framework to the different subsets of the whole dataset and provide better performance on each, we refrain from doing this as our focus is on generality across the subsets of images provided for the competition.
%
%The rest of the paper is organized as follows. Next section provides a review of the recent literature. Section \ref{approach} describes the complete deep learning framework. Section \ref{experiments} contains the detailed description of the experimental setup and results. Finally, we finish this paper with conclusive remarks and future research directions.

\section{Related Work}
\label{related-works}

We classify the recent literature performing leaf counting either directly or via instance segmentation into three categories, i.e. Leaf Segmentation Challenge in CVPPP-2014 (LSC-2014), Leaf Counting Challenge in CVPPP-2015 (LCC-2015), and others. Below we provide a brief account of the methods under each of these categories.

\textbf{LSC-2014:} In total, 4 methods evolve from this competition \cite{leaf_collation2016}. Although the training dataset for the competition included individual leaf instances indicated by different colors as the ground-truth, none of the 4 approaches use that ground truth to solve the instance segmentation problem. From that standpoint, they are all are eligible for the LCC-2017 competition also. The winner of this competition is IPK \cite{ipk2014, leaf_collation2016}. This method utilizes $3D$ histogram of the Lab color space of the training images to model both plant regions and background and test pixels are inferred non-parametrically using direct interpolation on the training data. Then, leaf centers are extracted using mathematical morphology of the distance map of the segmented foreground. These centers along with the foreground segmentation are processed by  heuristics-based graph algorithms to generate final instance segmentation map. Next, comes the unsupervised Nottingham approach \cite{leaf_collation2016}, which segments the foreground using seeded region growing \cite{seeded-region-growing} over the superpixels \cite{SLIC} extracted from the Lab color map. For the subsets of the dataset containing non-overlapping images, empirical thresholds are used instead of the superpixel means as the initial seed. Like IPK \cite{ipk2014}, they compute the distance map over the foreground pixels. Then, superpixels with centroids nearest to the local maxima in the distance map are chosen as the initial seeds with the assumption that they represent leaf centers the best for watershed based instance segmentation \cite{watershed-digital}. The MSU approach is adopted from the literature on multiple leaf alignment and tracking \cite{alignment_01, tracking_01, joint_seg_align_track} and primarily based on template matching based on Chamfer Matching algorithm \cite{chamfer_matching}. The authors use empirical threshold on the ``a" plane of the Lab image to select foreground candidates on which template matching is performed. The main drawbacks of this approach are manual selection of both the threshold and the templates and exhaustive template matching with a large number of templates, i.e. $1080$ templates for $2$ subsets and $1920$ for another. The last method submitted in LSC-2014 is Wageningen \cite{leaf_collation2016}. To segment the plant regions, the authors of this approach train a simple artificial neural network comprising one hidden layer of $10$ units with six pixel-based features, i.e. red ($R$), green ($G$), blue ($B$), excessive green ($2G-R-B$), and variance and gradient magnitude of filtered green pixel values, and then post-process the network output using morphological operations with heuristically chosen parameters. After that, watershed transform \cite{DIP_gonzalez} followed by empirical threshold based merging is performed to produce the instance segmentation result. A limitation of this method is the use of simple pixel features for foreground segmentation without using any contextual information in depth, resulting in the heavy usage of morphology to fine-tune the network output afterward.

\textbf{LCC-2015:} Only the winning method of LCC-2015 competition, General Leaf Counting (GLC) \cite{cvppp2015_winner}, is published in CVPPP-2015. To the best of our knowledge, this is the first approach posing the leaf counting problem as a nonlinear regression problem. The authors transform the original RGB image into a log-polar image \cite{log-polar} prior to further processing it to exploit the radial structure of the plants. Next, from the log-polar image, they extract patches based on the ground-truth foreground-background ratio in a sliding window fashion. These patch features are further vectorized with K-means \cite{bishop-ml} and triangle encoding \cite{triangle-encoding}. Lastly, max-pooling over the patch features is performed to form the final feature vector for each image and a support vector regression network \cite{cvppp2015_winner, support-vector-networks} is trained for the prediction task. A limitation of this system is that the authors use ground-truth plant segmentations in both training and testing phases of the counting module. While approximate plant segmentations could be generated by other methods~\cite{minervini2014image}, the study used perfect segmentations and therefore it is not clear how robust their counting module is to noisy or imperfect segmentations that are typical of automatic segmentation procedures.

\textbf{Others:} All the methods proposed since LCC-2015, addressing either the direct counting problem or counting by instance segmentation
%\cite{dpp-jubbens, ris2016, end-to-end_toronto_cvpr2017},
are found to be based on deep learning, which is not surprising given the resurgence of this subfield of machine learning in recent years. In the recurrent instance segmentation (RIS) approach \cite{ris2016}, the authors harness the power of sequential input processing of recurrent neural networks (RNN) \cite{rnn-graves} with the convolutional version of LSTM cells \cite{lstm} to segment out one leaf instance at a time. Unlike the use of LSTM and RNN in natural language processing, the idea is to use convolutional LSTM instead of the original formulation to facilitate the training of the network by mitigating the computational complexity of fully connected layers as well as exploiting the semi-global statistical properties of images. To deal with the problem of possible ordering of individual instances in the image, the authors formulate the loss function based on the relaxed version of intersection over union (IoU) \cite{relaxed-iou} and cross-entropy. The work done by Ren and Zemel \cite{end-to-end_toronto_cvpr2017} also use RNN similar to RIS \cite{ris2016}. However, their approach is primarily focused on extracting small patches each time to segment one instance using a similar idea of recurrent attention model \cite{ram-model} and then processing that small patch with LSTM \cite{lstm} and a deconvolutional network \cite{deconvnet} like architecture to segment a single instance. At the time of this writing, this work demonstrates the state-of-the-art performance for instance segmentation. Both this work and RIS use instance-level ground truth to train their networks and are therefore not directly comparable against ours. Nonetheless, we list their performance results in the \emph{Experiments} section. Finally, the deep plant phenomics (DPP) approach \cite{dpp-jubbens} proposes a method addressing the problem of counting directly without both plant segmentation and instance segmentation. The authors customize their architectures as well as input dimensions to achieve state-of-the-art accuracy on different subsets of the LCC-2015 dataset. However, it is not known if the approach would generalize and if a single DPP network would provide consistent results across all datasets.
%
% Ian - since each network is trained separately on each dataset, the background is the same
%
%The augmentation strategy employed in DPP uses random image cropping, which assumes images of centered and small rosettes.
Moreover, their training strategy relies on certain assumptions based on the nature of the images available in the LCC-2015 dataset; therefore, it is not clear how this approach would perform on the new types of images in the LCC-2017 competition. We will provide a detailed discussion about these issues while comparing our framework to DPP in section \ref{experiments}.

\begin{figure}[h]
	\centering
	\includegraphics[width=3.3in]{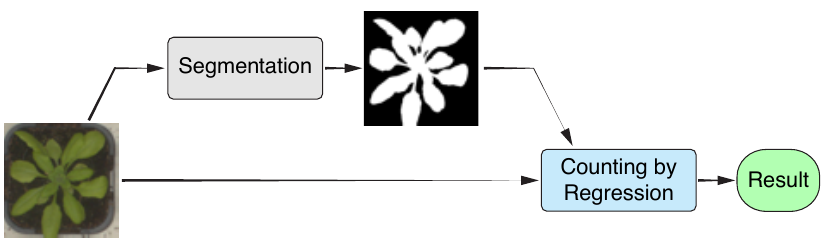}
	\caption{Block diagram of our approach.}
	\label{fig:block_diagram}
\end{figure}

\section{Our Approach}
\label{approach}
The approach presented in this section is developed to participate in the LCC competition \cite{lcc-2017} at CVPPP 2017. The high-level design of our framework follows a traditional computer vision workflow where the segmentation module is followed by the counting module (Figure \ref{fig:block_diagram}). Within each module, we incorporate task-specific convolutional architectures, which are trained without explicit knowledge of the plant species to develop a generalized framework able to learn only from the data. The architectures used for segmentation and counting are trained separately, but not independently since the binary mask generated by the segmentation model is used to train the counting model in conjunction with the RGB channels. In the following two subsections, we will describe the architectures along with the rationale behind their design. Training methodologies and data augmentation strategies for these models are described in the experiments section.

\begin{figure}[htbp]
	\centering
	\includegraphics[width=2cm, height=2cm]{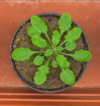}
	\includegraphics[width=2cm, height=2cm]{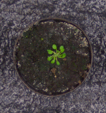}
	\includegraphics[width=2cm, height=2cm]{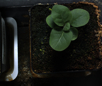}
	\includegraphics[width=2cm, height=2cm]{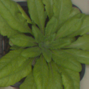}
    \caption{Sample images from the training set of CVPPP-2017 dataset \cite{lcc-2017, dataset-cvppp2017-01, dataset-cvppp2017-02, dataset-cvppp2017-03}. Representative images are taken and scaled from 4 training directories $A1$, $A2$, $A3$, and $A4$, respectively. }
    \label{fig:semi-global}
\end{figure}

\begin{figure*}[t]
	\centering
	\includegraphics[width=\textwidth]{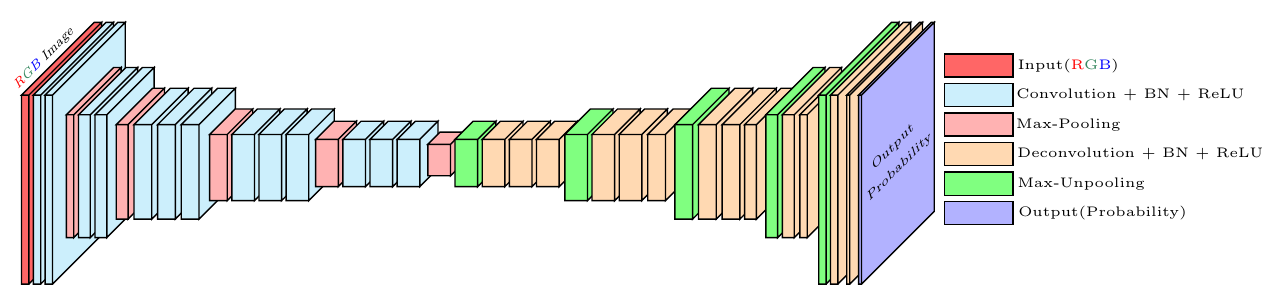}
    \caption{SegNet architecture \cite{segnet} used for leaf segmentation. Each of the convolution and deconvolution layers is followed by batch normalization (BN)~\cite{batch_normalization} and rectified linear unit (ReLU). All the pooling operations are $2\times 2$ max-pooling with stride of 2. Similarly, the unpooling operations are $2\times 2$ max-unpooling using the pooled indices taken from their corresponding max-pooling operations in the front-end of the network.}
    \label{fig:segnet}
\end{figure*}

\subsection{Segmentation}
The segmentation problem we address is that of differentiating the plant or foreground pixels, from the background. This kind of problem is also known as semantic segmentation where the semantics of the objects are utilized to accomplish the task. In recent years, many papers~\cite{fcn, deeplab, crf-rnn}
%\cite{fcn, segnet, deconvnet, deeplab, crf-rnn, review-semantic-segmentation}
 have been published addressing the solution for semantic segmentation from RGB images. Some of these architectures belong to the class of neural networks called \emph{deconvolutional networks} \cite{deconvnet, segnet}. The main idea behind this kind of network is to construct a compact and informative set of feature maps or vectors from a set of input images, and then generate class-probability maps from the feature maps. Like other convolutional networks, construction of the feature set from the input data is done by a convolutional sub-network comprising multiple layers of convolution, pooling, and normalization operations. This convolutional sub-network is followed by a deconvolutional sub-network consisting of convolution-transpose, unpooling, and normalization operations to generate the desired probability maps. From the standpoint of semantic segmenatation, both height and width of the input and the output are the same. Hence, the deconvolutional part of the network is designed as a mirrored version of the convolutional part, except the input and the output layers, irrespective of the complexity of the problem and the dimensionality of the class-space.

 Usually the design of a deconvolutional network contains fully connected (FC) layers in the middle to generate the feature vector from the pooled feature maps \cite{deconvnet}.
 %
 % fcn paper does not appear to use FC layers, so I've removed the citation here (also fcn doesn't appear to be dconvolutional either, just a conv net to get low res probability maps)
 %
 The FC layers are used to extract features in the global context for segmentation, and are therefore important if global context is necessary for the segmentation task. However, we propose that features in the semi-global context should be sufficient to segment the leaf regions from the background in color images, and therefore the FC layers could be omitted for our application. An advantage of eliminating the FC layers is that it considerably reduces the number of trainable parameters without sacrificing performance. For these reasons, we adopt the SegNet architecture \cite{segnet}, which omits FC layers and has shown promising results on SUN RGB-D \cite{sun-rgbd} dataset comprising complicated indoor scenes and CamVid \cite{camvid} video dataset of road scenes.
The removal of FC layers in SegNet results in about 90\% reduction of the number of trainable parameters as well as computational complexity. Figure \ref{fig:segnet} depicts the segmentation network we employ. The front-end convolutional sub-structure of the network is the VGG architecture \cite{vgg} with batch normalization followed by each convolutional layer. In the convolutional front-end of SegNet, there are five $2 \times 2$ pooling operations with zero overlapping following multiple convolution and rectification layers each time. Hence, the convolved feature maps are compressed $32$ times before starting the decompression via the deconvolutional back-end. We hypothesize that such level of compression or semi-global consideration is qualitatively sufficient to solve a comparatively easier problem of whole plant segmentation (Figure \ref{fig:semi-global}) as compared to other domains of semantic segmentation.

%\begin{figure}
%	\centering
%	\includegraphics[scale=0.441]{A1_plant009_rgb_resize.png}
%	\includegraphics[height=4.94cm,width=1.6cm]{A2_plant033_rgb_resize.png}
%	\includegraphics[height=4.94cm,width=1.6cm]{A3_plant029_rgb_resize.png}
%	\includegraphics[height=4.94cm,width=1.6cm]{A3_plant055_rgb_resize.png}
%	\includegraphics[height=4.94cm,width=1.6cm]{A5_plant077_rgb_resize.png} \\
%	\includegraphics[scale=0.44]{A1_plant122_rgb_resize.png}
%	\includegraphics[height=4.94cm,width=1.6cm]{A3_plant061_rgb_resize.png}
%	\includegraphics[height=4.94cm,width=1.6cm]{A5_plant083_rgb_resize.png}
%	\includegraphics[height=4.94cm,width=1.6cm]{A5_plant0865_rgb_resize.png}
%	\includegraphics[height=4.94cm,width=1.6cm]{A5_plant070_rgb_resize.png}
%	\caption{Sample images with corresponding binary segmentations. Rows $1$ and $4$ show the RGB images, rows $2$ and $5$ are the corresponding ground truth segmentations, and $3^{rd}$ and $6^{th}$ rows represent the segmentation results.}
%	\label{fig:result_seg}
%\end{figure}

\subsection{Counting}
\begin{figure} [!htbp]
	\centering
	\includegraphics[scale=0.441]{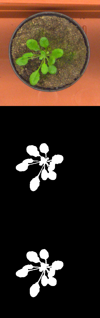}
	\includegraphics[height=4.94cm,width=1.6cm]{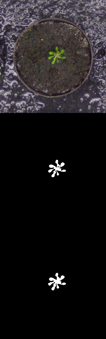}
	\includegraphics[height=4.94cm,width=1.6cm]{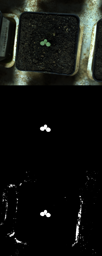}
	\includegraphics[height=4.94cm,width=1.6cm]{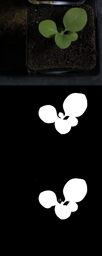}
	\includegraphics[height=4.94cm,width=1.6cm]{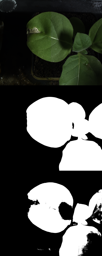}
	\caption{Sample images with corresponding binary segmentations: original RGB images (top row), corresponding ground truth segmentations (middle row), and our segmentation results generated by SegNet (bottom row).}
	\label{fig:result_seg}
\end{figure}

As shown in Figure \ref{fig:block_diagram}, we use both the RGB image and the corresponding binary segmentation image to estimate the number of leaves after the segmentation is done. The rationale behind providing the counting module with the segmentation mask and the original RGB image instead of providing either the segmented region in the RGB image or the binary mask alone will be evident from Figure \ref{fig:result_seg}. Although the segmentation results generated by SegNet are sufficiently accurate for the counting phase for many images in the dataset, our network generates spurious segmentations for few of them. The poorly segmented images generally have lower average intensities and regions of leaves where the color and texture properties are washed out or blurred. We expect our network to do more or less accurate segmentation for these images by using semi-global contextual information, but we believe it fails due to the low number of available samples of that kind in the training dataset both in terms of absolute count and ratio of the samples of this particular kind to other kinds. The problem of this data scarcity is specific to the data-hungry approaches like deep learning, which requires a substantial number of training instances of a particular prototype to generate an accurate input-output mapping for that specific type.

\begin{figure*}[t]
	\centering
	\includegraphics[scale=1.55]{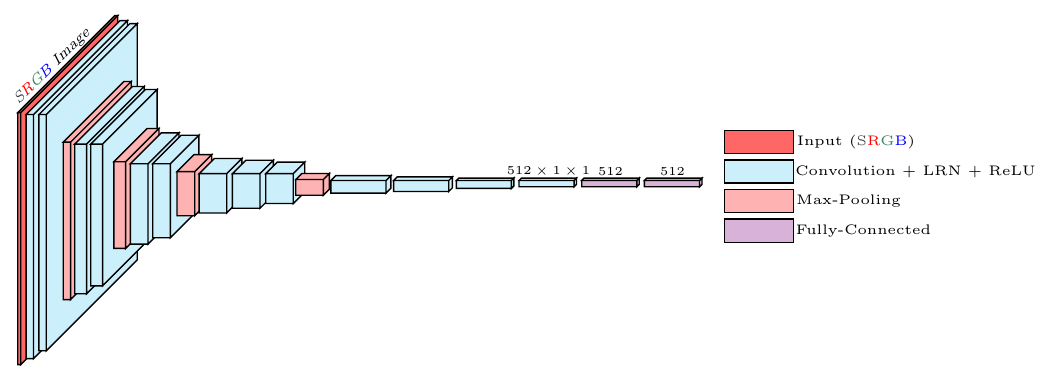}
    \caption{Counting architecture used for estimating the number of leaves from SRGB (Segmentation + RGB) channels. Each of the convolution blocks is a combination of convolution, local reponse normalization~\cite{alexnet}, and rectified linear unit (ReLU). All the pooling operations are $2\times 2$ max-pooling with stride of 2.}
    \label{fig:countnet}
\end{figure*}

Therefore, providing both the segmentation and the original image as input, we hope to influence the network to recover the missed plant regions as well as reject the false detections from original image with the help of segmentation mask for counting. We call this four channel input as the SRGB (Segmentation + RGB) image. We also expect that providing the segmentation channel as input to the leaf counting network will help to suppress bias from features in the background of the training images, such as the soil, moss, pot/tray color, which will vary between datasets.

%In other words, our counting by regression architecture shown in Figure \ref{fig:countnet} behaves like a heuristic search algorithm \cite{aima-book} in the first place, where it tries to extract the necessary regions from the original image using SRGB heuristics suitable for solving the counting or regression task afterwards.

The design of our leaf counting by regression network takes inspiration from the VGG architecture \cite{vgg}, which reinforces the idea of deeper architectures with a long list of convolutional and rectification layers stacked one after another with several pooling layers in between and then the classification layer follows a couple of fully connected layers. Usually, this kind of convolutional networks use suitable amount of padding to maintain fixed height and width of the feature maps. Padding the input maps serves well when the network is trained with large-scale datasets containing samples in the order of millions. However, in our case, we have a small dataset of several hundred images to train, which is very difficult to augment beyond several thousand images. Hence, to retain the power of deeper architecture and to train the parameters without significant overfitting at the same time, we reduce the number of parameters effectively by using convolution without padding throughout the network. Moreover, we choose the filter size of the convolutional layers in such a way that before proceeding through the fully connected layers, the feature map turns into a vector. Thus, with zero padding and careful choice of filter size, we are able to reduce the number of parameters from $49M$ to $30M$. Implementation details for both segmentation and regression networks are provided in the following section.

\section{Experiments}
\label{experiments}

In this section, we provide a detailed account of our experimental setup. First, we describe the dataset used for evaluation. Next, the training strategies for both networks are specified. Finally, the performance of our framework is analyzed and compared against state-of-the-art literature from both quantitative and qualitative standpoints.

\subsection{Dataset}
The dataset we use to evaluate our framework is provided to the teams registered for the Leaf Counting Challenge (LCC-2017). The objective of this challenge is to come up with the solutions able to count the number of leaves from plant images directly via learning algorithms without detecting individual leaf instances. All the RGB images in the dataset belong to either Tobacco or Arabidopsis plants. For the LCC competition, each RGB image is accompanied by a binary segmentation mask with $1$ and $0$ indicating plant and background pixels, respectively, and a center binary image with leaves centers denoted by single pixels.

The training dataset is organized into $4$ directories, namely $A1$, $A2$, $A3$, and $A4$. Directories $A1$ and $A2$ contain Arabidopsis images taken from growth chamber experiments with larger but different field of views covering many plants and then cropped to a single plant. Directory $A3$ enlists the Tobacco images with the field of view chosen to encompass a single plant. $A4$ is a subset of another public Arabidopsis dataset \cite{dataset-cvppp2017-03} collected using a time-lapse camera. In total, there are $27$ Tobacco images in $A3$, and $783$ Arabidopsis images in the rest of the directories. The organizers denote these directories along with the images as ``SPLIT" images since they are split into separate folders according to the origin. In addition, all these directories contain CSV files including ground truth leaf counts under the same nomenclature.

The ``SPLIT" directory structure for the testing set is the same as training, except that it includes an extra directory denoted by $A5$, enlisting images from different sources of origin altogether with the objective to emulate a leaf counting task in the wild. Hence, the organizers represent $A5$ images under the nomenclature ``WILD".

\subsection{Training and Implementation}

\textbf{SegNet training:} Unlike training in the original SegNet paper \cite{segnet}, we trained our model from scratch without using any pretrained weights for initialization.  Also in SegNet, the authors used different learning rates for different modules, whereas a fixed learning rate was used for all the layers in our training.

We used an input and output image size of $224\times 224$ pixels in SegNet, whereas the original image size was approximately $500\times 500$ and $2000\times 2500$. While training deeper networks, the obvious advantage of using smaller input-output size than the original ones is data augmentation up to a considerable amount. We augmented the data and train the network in $3$ stages. First, for each image, we extracted the union of top 20 object proposals \cite{edge-boxes}, flipped top-bottom and left-right, rotated them with an angular step size of 4 degree, cropped the largest square from the center position to avoid dark regions due to rotation, and created a couple of Gaussian blurred version and corresponding sharpened images. In this way, we generated about $0.8M$ augmented samples from $810$ original images and trained the network for 5 epochs with randomly cropped $224\times 224$ subsamples. Second, we took the proposal images and their flipped versions and generated nearly $0.3M$ subsamples of size $224 \times 224$ deterministically with a fixed stride and train the network for another 8 epochs. Finally, we generated another $0.19M$ samples in a similar manner as in the second step, but this time from the original images instead of the proposals. Then, we fine-tuned the network with these $0.19M$ samples for 37 epochs. In all stages, SGD-momentum was used as the optimizer with initial learning rate, momentum and weight decay of $0.01$, $0.9$, and $0.0001$, respectively and these parameters were changed later based on the training statistics. Spatial cross-entropy was used as the error criterion. The ratios of foreground to background weights in the cross-entropy calculation for the first stage was $2.0$ and $1.2$ for the later steps. In the test phase, we took dense $224\times 224$ samples deterministically with fixed stride from each of the test images and classified each pixel based on the aggregate probability over the samples. We initialized the convolutional weights with Xavier \cite{xavier} prior to the start of training.

\begin{figure}[h]
\centering
	\includegraphics[scale=0.40]{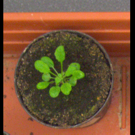}
	\includegraphics[scale=0.40]{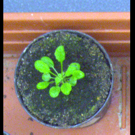}
	\includegraphics[scale=0.40]{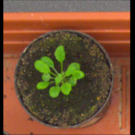} \\
	\includegraphics[scale=0.40]{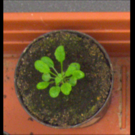}
	\includegraphics[scale=0.40]{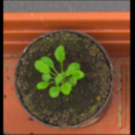}
	\includegraphics[scale=0.40]{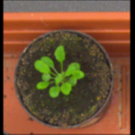} \\
	\includegraphics[scale=0.40]{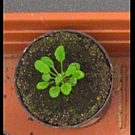}
	\includegraphics[scale=0.40]{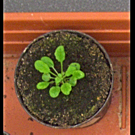}
	\includegraphics[scale=0.40]{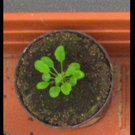}
	\caption{Augmentation samples for training the counting network.}
	\label{fig:aug_samples}
\end{figure}

\begin{table*}[!htbp]
\centering
\caption{Head-to-head comparison against LCC-2015 winner. Note that \emph{All} refers to A1-A3 for GLC and A1-A5 for Ours.}
\label{tab:lcc-comparison}
\begin{adjustbox}{scale=1.0}
\begin{tabular}{c|cc|cc|cc|cc}
\hline
\multirow{2}{*}{Directories} & \multicolumn{2}{c|}{CountDiff} & \multicolumn{2}{c|}{AbsCountDiff} & \multicolumn{2}{c|}{PercentAgreement {[}\%{]}} & \multicolumn{2}{c}{MSE} \\ \cline{2-9}
 & GLC\cite{cvppp2015_winner} & Ours & GLC\cite{cvppp2015_winner} & Ours & GLC\cite{cvppp2015_winner} & Ours & GLC\cite{cvppp2015_winner} & Ours \\ \hline
A1 & -0.79(1.54) & -0.33(1.38) & 1.27(1.15) & 1.00(1.00) & 27.3 & 30.3 & 2.91 & 1.97 \\
A2 & -2.44(2.88) & -0.22(1.86) & 2.44(2.88) & 1.56(0.88) & 44.4 & 11.1 & 13.33 & 3.11 \\
A3 & -0.04(1.93) & \textcolor{red}{2.71(4.58)} & 1.36(1.37) & \textcolor{red}{3.46(4.04)} & 19.6 & \textcolor{red}{7.1} & 3.68 & \textcolor{red}{28.00} \\ \hline
A4 & - & 0.23(1.44) & - & 1.08(0.97) & - & 29.2 & - & 2.11 \\
A5 & - & 0.80(2.77) & - & 1.66(2.36) & - & 23.8 & - & 8.28 \\ \hline
\emph{All} & -0.51(2.02) & 0.73(2.72) & 1.43(1.51) & 1.62(2.30) & 24.5 & 24.0 & 4.31 & 7.90 \\ \hline
\end{tabular}
\end{adjustbox}
\end{table*}

\begin{table*}[!htbp]
\centering
\caption{Comparison against state-of-the-art literature. Note that \emph{All} refers to A1-A3 for previous work and A1-A5 for Ours.}
\label{tab:full-comparison}
\setlength\tabcolsep{1.5pt}
\begin{adjustbox}{max width=\textwidth}
\begin{tabular}{|l|c|c|c|c|c|c|c|c|c|c|c|c|}
\hline
\multicolumn{1}{|c|}{\multirow{2}{*}{Methods}} & \multicolumn{6}{c|}{CountDiff} & \multicolumn{6}{c|}{AbsCountDiff} \\ \cline{2-13}
\multicolumn{1}{|c|}{} & A1 & A2 & A3 & A4 & A5 & \emph{All} & A1 & A2 & A3 & A4 & A5 & \emph{All} \\ \hline
IPK\cite{ipk2014} & -1.8(1.8) & -1.0(1.5) & -2.0(3.2) & - & - & -1.9(2.7) & 2.2(1.3) & 1.2(1.3) & 2.8(2.5) & - & - & 2.4(2.1) \\
Nottingham\cite{leaf_collation2016} & -3.5(2.4) & -1.9(1.7) & -1.9(2.9) & - & - & -2.4(2.8) & 3.8(1.9) & 1.9(1.7) & 2.5(2.4) & - & - & 2.9(2.3) \\
MSU\cite{leaf_collation2016} & -2.5(1.5) & -2.0(1.5) & -2.3(1.9) & - & - & -2.3(1.8) & 2.5(1.5) & 2.0(1.5) & 2.3(1.9) & - & - & 2.4(1.7) \\
Wageningen\cite{leaf_collation2016} & 1.3(2.4) & -0.2(0.7) & 1.8(5.5) & - & - & 1.5(4.4) & 2.2(1.6) & 0.4(0.5) & 3.0(4.9) & - & - & 2.5(3.9) \\
GLC\cite{cvppp2015_winner} & -0.79(1.54) & -2.44(2.88) & -0.04(1.93) & - & - & -0.51(2.02) & 1.27(1.15) & 2.44(2.88) & 1.36(1.37) & - & - & 1.43(1.51) \\
DPP\cite{dpp-jubbens} & - & - & - & - & - & - & 0.41(0.44) & 0.61(0.47) & 0.61(0.54) & - & - & - \\
RIS+CRF\cite{ris2016} & - & - & - & - & - & 0.2(1.4) & - & - & - & - & - & 1.1(0.9) \\
EERA\cite{end-to-end_toronto_cvpr2017} & - & - & - & - & - & - & - & - & - & - & - & 0.8(1.0) \\
Ours & -0.33(1.38) & -0.22(1.86) & 2.71(4.58) & 0.23(1.44) & 0.80(2.77) & 0.73(2.72) & 1.00(1.00) & 1.56(0.88) & 3.46(4.04) & 1.08(0.97) & 1.66(2.36) & 1.62(2.30) \\ \hline
\end{tabular}
\end{adjustbox}
\end{table*}

\begin{table*}[!htbp]
\centering
\caption{Possible interpretation of the performance measures.}
\label{tab:interpretation}
\setlength\tabcolsep{1.5pt}
\begin{adjustbox}{max width=\textwidth}
\begin{tabular}{|l|l|}
\hline
Measures & Possible Interpretation \\ \hline
CountDiff $\downarrow$ & The model is less biased towards overestimate or underestimate. \\ \hline
AbsCountDiff $\downarrow$ & Average performance is better. \\ \hline
PercentAgreement $\uparrow$ & Number of accurate predictions is higher. \\ \hline
CountDiff $\downarrow$, AbsCountDiff $\downarrow$ & Less bias with better performance. Desirable properties of an ideal nonlinear regression model. \\ \hline
CountDiff $\downarrow$,  AbsCountDiff $\uparrow$ & High positive and negative errors cancel out. Model behaviour tends to be linear than usual. \\ \hline
PercentAgreement $\downarrow$, AbsCountDiff $\downarrow$ & \begin{tabular}[c]{@{}l@{}} Although many predictions are not exactly accurate, all of the predictions are close to the original; \\ therefore, model performance is uniform over the samples.\end{tabular} \\ \hline
PercentAgreement $\uparrow$, AbsCountDiff $\uparrow$ & \begin{tabular}[c]{@{}l@{}} Although many predictions are exact, wrong predictions are far from the original; \\ therefore, model performance is not uniform over the samples.\end{tabular} \\ \hline
\end{tabular}
\end{adjustbox}
\label{tab-interp}
\end{table*}

\textbf{Count network training:} Training of the counting network is fairly straightforward compared to SegNet. In this phase, we used all the images as a whole without prior cropping or sampling operation for data augmentation to ensure that the ground truth leaf counts were valid for all augmented images. Also, while designing the network architecture, we experimented with adaptive operations to deal with variable sized images, but they did not seem to work better than resizing the images to a fixed size. Moreover, we had to be cautious in the choice of the size for resizing operation so that for bigger images with resolution like $2000 \times 2500$, properties of the small leaf regions did not deteriorate much. Considering this fact, we chose the modified image size to be $448 \times 448$ preserving the aspect ratio. Thus, the largest dimension was taken to be $448$ and the smaller one was padded with zeros afterward.

After the resize operation was performed, each of the images was augmented $8$ times using intensity saturation, Gaussian blurring and sharpening, and additive Gaussian noise (Figure \ref{fig:aug_samples}). Each image was also flipped top-bottom and left-right and rotated 180\si{\degree} along with similar augmentations. Thus, we generated $36$ slightly different samples with the same ground truth from each original image, resulting in $29160$ training instances for the regression network.

After the data generation was done, the counting or regression network was trained for 40 epochs using Adam \cite{adam} with fixed learning rate and weight decay both set to $0.0001$. $\text{Smooth-}L_1$ criterion was used as the loss function instead of simple $L_1$ criterion to prevent gradient explosions as described in \cite{fast-rcnn}. At first, we started training the model with normalized FC layers of size $1024$. However, based upon the training statistics and to reduce the risk of overfitting, we changed the size of FC layers to $512$ and retrained the model with the already trained convolutional weights. Finally, the model trained until epoch $35$ was used to generate the prediction for final submission.

\textbf{Implementation:} We used Torch\cite{torch} as the deep learning framework for both models. All the convolutional filters of the segmentation network were of size $3\times 3$. For regression architecture, $9\times 9$ convolution was performed until the second max-pooling operation and $5\times5$ afterward. We used the convolutional stride of $1$ throughout both networks. All the pooling operations were $2\times 2$ max-pooling with stride of $2$. The dimension of all fully connected layers in the regression network was $512$. Training was performed on a single NVIDIA Quadro P6000 Dell workstation. On this machine, training of SegNet took about $6-7$ days, whereas the regression network was trained within a couple of days. Code is publicly available here. \footnote{\url{https://github.com/p2irc/leaf_count_ICCVW-2017}}

\subsection{Evaluation}

Evaluation of our complete framework was accomplished in three stages. First, we assessed the segmentation network in terms of the precision and recall (equation \ref{eq:prec_rec}) of the plant pixels. Next, we performed a head-to-head comparison against the winner of the previous LCC competition. Finally, we compared our results to the state-of-the-art approaches. We also performed an ablation study by training our counting network with and without the segmentation channel as input, in order to cast some light on the issue regarding the need for foreground segmentation.

\textbf{Foreground segmentation:} Even though the accuracy of binary segmentation is not a criterion for evaluation in the LCC competition \cite{lcc-2017}, we provide precision and recall (equation \ref{eq:prec_rec}) of our segmentation model in Table \ref{tab:binseg} to justify our assumption on the sufficiency of semi-global context for leaf segmentation. It is evident from Table \ref{tab:binseg} that the segmentation results generated by SegNet using semi-global information are good enough to be used for the regression network. Performance of the segmentation network is comparatively lower for directory $A3$ (Table \ref{tab:binseg}, red text) since there are only $27$ Tobacco images in the $A3$ training set as compared to $783$ Arabidopsis images in the rest of the directories.

\begin{equation}
\begin{cases}
	\text{Precision} = \frac{\text{True Positive}}{\text{True Positive} \, + \, \text{False Positive} } \\
	\text{Recall} = \frac{\text{True Positive}}{\text{True Positive} \, + \, \text{False Negative}}
\end{cases}
\label{eq:prec_rec}
\end{equation}

\begin{table}[!htbp]
\centering
\caption{Binary segmentation results.}
\label{tab:binseg}
\begin{tabular}{|l|l|l|l|l|l|}
\hline
Directory & A1 & A2 & A3 & A4 & A5 \\ \hline
Precision & 0.98 & 0.94 & \textcolor{red}{0.80} & 0.96 & 0.92 \\ \hline
Recall & 0.99 & 0.99 & 0.94 & 0.98 & 0.97 \\ \hline
\end{tabular}
\end{table}

\textbf{Comparison against the previous winner:} Next, we provide comparisons in both Table \ref{tab:lcc-comparison} and \ref{tab:full-comparison}. Table \ref{tab:full-comparison} provides comparisons against all the recent literature, whereas Table \ref{tab:lcc-comparison} provides a head-to-head comparison against the LCC-2015 winner, which is more detailed due to the availability of the performance metrics for \cite{cvppp2015_winner}. In Table \ref{tab:lcc-comparison}, ``CountDiff" refers to the mean and standard deviation (shown in parentheses) of the difference in count averaged over images. ``AbsCountDiff" is the absolute of ``CountDiff". The term ``PercentAgreement" indicates the percentage of exact matches between the actual prediction and ground truth measurement for counts. ``MSE" is the abbreviation for mean-squared error.
%Table~\ref{tab-interp} provides a summary of our interpretation of different scenarios of the performance measures.

From Table \ref{tab:lcc-comparison}, it is evident that we achieve lower CountDiff and AbsCountDiff for directories $A1$ and $A2$. Lower CountDiff means that our model is less biased towards underestimation or overestimation than GLC \cite{cvppp2015_winner}, whereas lower AbsCountDiff can be interpreted as the indicator of better average performance of the system. However, our framework performs poorly on directory $A3$ (Table \ref{tab:lcc-comparison}, red text). The reason behind the failure is pretty straightforward. Note that, in the training set, there are in total $783$ Arabidopsis images in $A1$, $A2$, and $A4$. On the other hand, there are only $27$ Tobacco images in $A3$, which is scarce for the types of deep architectures we are using that contain millions of parameters. Hence, our regression network fails to model the distribution for leaf counting over the Tobacco images. This inadequacy is also reflected in the AbsCountDiff measure for the test directory $A5$, which is a mixture of Arabidopsis and Tobacco images altogether.

For directory $A2$, although our CountDiff and AbsCountDiff are better than those of GLC, PercentAgreement of GLC is much better than ours. Apparently, it might seem to be a pitfall of our system. However, the combination of lower AbsCountDiff and lower PercentAgreement means that even though the number of exact predictions is low, all the predictions are pretty close to the original and the overall performance of the system is more or less uniform over the test images. On the contrary, comparatively higher values of AbsCountDiff and PercentAgreement, which belong to GLC for directory $A2$, refers to the situation where model performance is not uniform over the samples. In other words, predictions may be accurate for easier samples with no leaf overlap or moderate-sized leaves or both, but deteriorate for harder cases with smaller or overlapping leaves. In that sense, our generalized framework is capable of modeling and inferring leaf shapes under deformation and partial occlusion better than GLC given a few hundred images for a particular species. To facilitate this kind of comparative evaluation of our method by the readers, we enlist a set of combinations of the measures along with their possible interpretations in Table \ref{tab:interpretation}. Also, note that our average measurement (directory ``\emph{All}") is over $501$ test images from $5$ directories (A1-A5), whereas the average for GLC is taken over $98$ test images from $3$ directories (A1-A3).

\textbf{General comparison:} Table \ref{tab:full-comparison} shows that our method performs well as compared to all the LSC-2014 \cite{ipk2014, leaf_collation2016} and LCC-2015 \cite{cvppp2015_winner}, except for the failure on directory $A3$ due to inadequate number of samples. Both RIS+CRF \cite{ris2016} and EERA \cite{end-to-end_toronto_cvpr2017} use instance-level ground truth. Hence, they are eligible for the segmentation competition (LSC), but not the counting competition (LCC). Nonetheless, we put them in the list to demonstrate our comparability to these state-of-the-art methods developed with instance segmentations that are more expensive in terms of training complexity/time and ground truth data requirements. DPP \cite{dpp-jubbens} is the only method close to ours in the style of approach, except that they use three shallow regression networks, each one highly customized over a single directory. Moreover, DPP uses random cropping from $10\% - 25\%$ for the purpose of data augmentation while training. This could result in mislabeled images if leaves are cropped out of certain images. The new rosette images in the LCC-2017 dataset include larger rosettes that cover more of the image frame (and extend outside the frame in certain cases, see rightmost image in Figure~\ref{fig:semi-global}); therefore it is not clear how DPP would perform on the larger and more varied test images in the new competition dataset.

%This implies that for each randomly cropped sub-image, they use the same ground truth as the original full-sized image with the assumption that it does not make much of a difference. Albeit their method seems to work well for this particular dataset where for most images, there is no leaf in the border region, this is a highly strict assumption merely based on the manual observation on that particular dataset. Also, at test time, they average their result over randomly cropped sub-images. It is highly likely that accidentally that the mean of a set of overestimates and underestimates gets closer to the original count at test time since no detailed analysis is provided on any of these facts without mere ``AbsCountDiff" on only $98$ test images under $A1$, $A2$, and $A3$ directories.

\textbf{Ablation study:} To justify the inclusion of a segmentation network within our framework, we performed an ablation study by training our regression network using only RGB images as the input without foreground segmentation. We found slower convergence than that of using the segmentation images as input. However, counting results using only RGB images were comparable, which supports the approach proposed by DPP of using a regression network directly on RGB images and that the network learns relevant features directly without a priori segmentation. Nonetheless, we do expect that providing foreground segmentation as an additional input channel helps to push the regression architecture to train on localized features within the plant region in the image. This might help to suppress background features that could limit the generalizability of the counting model if provided images of rosettes grown in different backgrounds, e.g. in different pots, trays, or growth tables. The issue of localization of features in these types of regression networks requires additional attention as future work.

\section*{Acknowledgment}
This work was undertaken thanks in part to funding from the Canada First Research Excellence Fund and the Natural Sciences and Engineering Research Council of Canada.

\section{Conclusion and Future Work}
\label{conclusion}
In this paper, as a participant of the LCC-2017 competition, we provide a complete and generalized data-driven framework for leaf counting from RGB images directly without instance segmentation. We demonstrate that given a moderate amount of data on any species, our architectures are able to learn to estimate the number of leaves without prior knowledge on that particular species or surroundings of the plant. From the perspective of informed search strategies, we do plant segmentation prior to counting with the assumption that the additional foreground segmentation channel guides the regression model to extract necessary features only from the plant region and thus trains the model correctly. However, based upon other recent works and ours, the need for segmentation prior to counting by the deep networks is still an open question. As future work, we plan to investigate this issue in more detail, with the goal of achieving equivalent performance to that of instance segmentation architectures with much simpler and easier to train non-recurrent networks such as reported in the present study.

{\small
\balance
\bibliographystyle{ieee}
\bibliography{egbib}
}

\end{document}